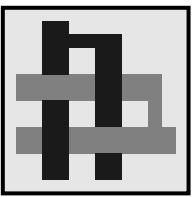

המעבדה לבלשנות חישובית
Laboratory for Computational Linguistics

cmp-lg/9407014  17 Jul 94

# Abstract Machine for Typed Feature Structures

by
Shuly Wintner and Nissim Francez

Technical Report #LCL 94-8
July 1994

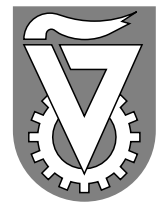

הטכניון – מכון טכנולוגי לישראל, חיפה 32000 ישראל
Technion - Israel Institute of Technology, Haifa 32000 Israel

# Abstract Machine for Typed Feature Structures

Shuly Wintner    Nissim Francez

July 13, 1994

**Abstract**

This paper describes a first step towards the definition of an abstract machine for linguistic formalisms that are based on typed feature structures, such as HPSG. The core design of the abstract machine is given in detail, including the compilation process from a high-level specification language to the abstract machine language and the implementation of the abstract instructions. We thus apply methods that were proved useful in computer science to the study of natural languages: a grammar specified using the formalism is endowed with an operational semantics. Currently, our machine supports the unification of simple feature structures, unification of sequences of such structures, cyclic structures and disjunction.

# 1   Introduction

Typed feature structures (TFSs) serve as a means for the specification of linguistic information in current linguistic formalisms such as HPSG or Categorial Grammar ([19, 20, 15]). Seen as a generalization of first-order terms (FOTs), TFSs are also used to specify logic programs and constraint systems in frameworks such as LOGIN ([4]), LIFE ([2]), ALE ([10, 8]), CUF ([11]), TFS ([26]) and others. Many general frameworks that are completely independent of any linguistic theory can be used to specify grammars for natural languages. Indeed, most of the above mentioned languages were used for specifying HPSG grammars. Different systems employ different kinds of TFSs, based on a variety of algebraic definitions, and we usually follow the representation of [9] here.

Linguistic formalisms (in particular, HPSG) use TFSs as the basic blocks for representing linguistic data: lexical items, phrases and rules. They usually do not specify a mechanism for manipulating TFSs: parsing algorithms, for instance, are external to the formalism. Moreover, HPSG has no formal definition yet, so that it is not fully determined what the characteristics of the formalism are, nor what the properties of a specific HPSG grammar are. In general constraint solvers based on TFSs the operations performed on the structures are more explicit, though such systems usually suffer severe efficiency problems. When no processing direction is specified, and the system searches the complete space of solutions for some specification, its performance is disappointing. Clearly, efficient processing calls for a different method.

In this paper we present a first step of an approach for processing TFSs that guarantees both an explicit definition and high efficiency. Viewing grammars for natural languages as formal specifications in a high-level programming language, we incorporate techniques that were proved valuable in computer science, especially in programming languages semantics, thus utilizing the benefits of bringing together the two paradigms: computer science and linguistics. Our main aim is to provide an operational semantics for TFS-based linguistic formalisms, especially HPSG. We adopt an abstract machine approach for the compilation of specifications of such grammars. The abstract machine comprises data structures and instructions, augmented by a compiler from the TFS formalism to the abstract instructions.

In designing the machine we try to capture the intuitive meaning of the linguistic formalism and to reflect it in the machine architecture. The operational semantics of each instruction is defined using a low-level language that can be executed on ordinary hardware. We thus expect a substantial improvement over existing parsers in both space and time requirements. Recently, a similar approach was applied to the LIFE language ([3]); however, due to differences in the motivation and in the formalisms, our machine is much different. As far as we know, this is the first attempt to use abstract machine for a linguistically motivated formalism.

The use of an abstract machine ensures that every grammar specified using our system is endowed with a concrete, well defined operational meaning. We thus provide a means for rigorously stating mathematical properties of specific grammars as well as entire formalisms. For example, it will enable to formally verify the correctness of a compiler for HPSG, given an independent definition; or to prove the equivalence of two HPSG grammars.

## 1.1 Related Works

The first language to combine feature structures, typing hierarchies and constraint specification is probably LOGIN ([4]). In this system FOTs are replaced by the more general $\psi$-*terms* and a partially-ordered set of types introduces a built-in inheritance to the language. A natural descent of LOGIN is LIFE ([2, 5]), where styles from functional programming are added to the basic constructs. An abstract machine for LIFE is currently under development and preliminary results were very recently described in [3].

Another effort, motivated by linguistic needs, is TFS ([26]); it uses a partially ordered set of types and allows general constraints over typed feature structures to be specified. An abstract rewrite machine is employed to resolve these constraints in no a-priory order, i.e., grammars can be used for both parsing and generation. This leads to very inefficient processing.

Recently two other systems were constructed that allow the specification of logical constraints over typed feature structures. ALE ([8]) is meant to be a general logic engine, where definite clauses over typed feature structures can be specified. In addition, the language provides means for specifying phrase structure rules, and a parsing algorithm is embedded within it. However, ALE limits the type hierarchy to be a bounded complete partial order, where unique unifiers exist for every consistent subset of types.

CUF ([11]) is more ambitious than the former systems as it is aimed to cover as many as possible of the extensions to simple unification formalisms. It supports partial ordering of types but doesn't demand it; feature structures can contain disjunction, negation, list- and set-values and arbitrary functional and relational constraints. Separate control statements guide the resolution process.

What is common to all the above-mentioned systems is that even though they can be used to specify natural language grammars, and indeed many of them were motivated by linguistic applications, they are very general and completely independent of any particular linguistic theory or formalism. Many of these frameworks were used for devising grammars for HPSG, but it is hard to compare the grammars that were designed for the different systems.

## 1.2 Abstract Machines

Abstract machines were used for various kinds of languages: starting from Landin's SECD ([18]), many compilers for functional languages were designed this way. Even imperative languages such as Pascal were implemented using abstract machine techniques (P-Code). When logic programming languages appeared, such techniques were applied to them as well. Notably, Warren designed an abstract machine for the execution of Prolog ([25], [1]). Even though there were prior interpreters and compilers for Prolog, it was the Warren Abstract Machine (WAM) that gave the language not only a good, efficient compiler, but, perhaps more importantly, an elegant operational semantics.

The WAM implementation of Prolog consists of a machine, augmented by a compiler into its instruction set. The meaning of each instruction is defined using a low-level language that can be mapped to any ordinary hardware. In fact, there is even a formal verification of the correctness of this implementation ([21]).

The WAM immediately became the starting point for many Prolog compilers. The techniques it delineates serve not only for Prolog proper, but also for constructing compilers for related languages. To list just a few examples, abstract machine techniques were used for a parallel Prolog compiler ([16]), for variants of Prolog that use different resolution methods ([24]), and for a general theorem prover ([22]).

A careful design of such an abstract architecture must compromise between two, usually conflicting, requirements: it must be close enough to the high-level language in order to capture its semantics and to accommodate simple compilation; on the other hand, it must remain close to common architectures so that its language can be efficiently executed.

## 1.3 Structure of the Document

The next section sketches the fundamental notions needed for understanding our design. We define type hierarchies, feature structures, well-typedness conditions and unification. We also give a FOT-like representation of TFSs. Then we extend the definitions to license sequences of TFSs. Section 3 describes the characteristics of the basic abstract machine we design: feature structure representation in memory, flattening an FOT-like TFS and simple unification of two TFSs. We also detail the process of compiling the type hierarchy. Section 4 presents extensions of this basic abstract machine: in section 4.1 we extend the machine so that it can handle sequences of feature structures; section 4.2 introduces disjunctive TFSs and delineates the changes in the design that are needed in order to manipulate them. A conclusion and plans for further research are given in section 5. Appendix A provides a more detailed mathematical background, in the lines of [9]. Appendix B lists all the machine instructions and auxiliary functions.

## 2 The Framework

### 2.1 Fundamental Notions

An HPSG grammar consists of a type specification (the **signature**) and grammar rules (including principles and lexical rules). The basic entity of HPSG is the **feature structure** which is a connected, directed, labeled, possibly cyclic, finite graph, whose nodes are decorated with **types** and whose edges are labeled by **features**. The types are ordered according to an **inheritance hierarchy** where higher types inherit features from their super-types. A formal definition of types and feature structures is given in Appendix A.1.

As there are many different formalizations of TFS systems it is important to define the framework with which we work; it is, with slight modifications, Carpenter's system ([8, 10]). First, we use a set of types that includes both $\bot$, the least type, and $\top$, the greatest one. We order types by **subsumption** according to their information content, not according to the cardinality of the set of elements they can be assigned to. This means that the type $\bot$ is the most general type, subsuming every other, and the type $\top$ is the contradictory type, subsumed by every other. In this we follow, e.g., [10, 9] but not [5, 23].

Second, we require that the type inheritance hierarchy be bounded complete, that is, every set of consistent types must have a unique least upper bound (other than $\top$). We also require that the appropriateness specification of the features and the types be such that every feature is introduced by some least type (and is appropriate for all the types it subsumes), and that appropriateness be monotone. A last requirement of the appropriateness specification is that it does not contain loops. For a formal presentation of the above notions see Appendix A.1.

Third, we require that the feature structures with which we deal be totally well-typed. This property is most convenient and results in more efficient processing. It is more problematic from the user's point of view, as users might find it useful to specify only partial information about linguistic entities. Therefore, some description language must be provided, such that the user is able to give partial descriptions from which totally well-typed feature structures can be automatically deduced. Various description languages were suggested for feature structures in general and TFS in particular. In many cases the manipulation of the structures (e.g., unification) is defined for the description rather than over the objects themselves. As there are efficient algorithms to deduce structures from their descriptions, we prefer not to commit ourselves to one description language. We define our system over explicit representations of TFS, as will be clear from section 2.3.

## 2.2 Type Specification

The first part of a program (or a grammar) is a type specification. As described above this should contain the type hierarchy and the appropriateness specification. We adopt Carpenter's format ([8]) for this specification: it is a sequence of statements of the form

$$t \text{ sub } [t_1, t_2, \ldots, t_n] \text{ intro } [f_1 : r_1, \ldots, f_m : r_m].$$

where $t, t_1, \ldots, t_n, r_1, \ldots, r_m$ are types, $f_1, \ldots, f_m$ are feature names and $n, m \geq 0$.

Such a statement, which is said to **characterize** $t$, means that $t_1, \ldots, t_n$ are subtypes of $t$ (i.e., for every $i, 1 \leq i \leq n, t \sqsubseteq t_i$), and that $t$ has the features $f_1, \ldots, f_m$ appropriate for it. Moreover, these features are **introduced** by $t$, i.e., they are not appropriate for any type $t'$ such that $t' \sqsubset t$. Finally, the statement specifies that the appropriate values for each feature $f_i$ in $t$ should be of type $r_i$. We demand that each type (except $\top$ and $\bot$) is characterized by exactly one statement.

The full subsumption hierarchy of the types is the reflexive transitive closure of the $\sqsubseteq$ relation as specified by the characterization statements. If this relation is not a bounded complete partial order, the specification is rendered invalid. The same is true in case it is not an appropriateness specification (see Definition A.3) or contains a loop (see Definition A.6).

We use the following type hierarchy $H$ as a running example:

```
bot sub [a,b,d].
  a sub [c] intro [f1:bot,f3:d1].
    c sub [] intro [f4:bot].
  b sub [c,e] intro [f2:bot,f3:d].
  d sub [d1,d2].
    d1 sub [].
    d2 sub [].
```

The type bot stands for $\bot$ in this specification. The type $\top$ is systematically omitted from type specifications.

The type-unification (or least upper bound) table, consisting of an entry for every pair of types, can be computed at compilation time. The appropriate table for the hierarchy $H$ is depicted[1] in figure 1. Note that the table encodes not only the features of the unified type, but also the 'origin' of these features. For example, the table entry for the types *a(f1,f3)* and *b(f2,f3)* is *c(f1,f2,f3,f4)*. This entry states that the first feature of the type $c$ is inherited from the type $a$, while the second is inherited from $b$. The third feature, $f3$, is common to both $a$ and $b$, while the last feature is a new one, introduced by $c$. In our representation of types we always assume that the features are totally ordered and that their order is given as part of the specification. In the examples below we assume that the order is the lexicographic one.

---
[1] As such tables are always symmetric, only their upper part is shown.

| unify_type(X,Y) | $\bot$ | a(f1,f3) | b(f2,f3) | c(f1,f2,f3,f4) | d | d1 | d2 | e(f2,f3) |
|---|---|---|---|---|---|---|---|---|
| $\bot$ | $\bot$ | a(f1,f3) | b(f2,f3) | c(f1,f2,f3,f4) | d | d1 | d2 | e(f2,f3) |
| a(f1,f3) | | a(f1,f3) | c(f1,f2,f3,f4) | c(f1,f2,f3,f4) | $\top$ | $\top$ | $\top$ | $\top$ |
| b(f2,f3) | | | b(f2,f3) | c(f1,f2,f3,f4) | $\top$ | $\top$ | $\top$ | e(f2,f3) |
| c(f1,f2,f3,f4) | | | | c(f1,f2,f3,f4) | $\top$ | $\top$ | $\top$ | $\top$ |
| d | | | | | d | d1 | d2 | $\top$ |
| d1 | | | | | | d1 | $\top$ | $\top$ |
| d2 | | | | | | | d2 | $\top$ |
| e(f2,f3) | | | | | | | | e(f2,f3) |

Figure 1: The type-unification table for $H$

## 2.3 Representation of Feature Structures

The most convenient graphical representation of feature structures is attribute-value matrices (AVMs). However, to represent a (totally well-typed) feature structure linearly we use a notation that resembles a first-order term, where the type plays a similar role to that of a function symbol and the features are listed in a fixed order. Reentrancy is implied by attaching identical tags to reentrant feature structures. This representation is based upon Aït-Kaci's $\psi$-terms ([4, 6]); its definition and semantics are given in Appendix A.2.

Total well-typedness means that the names of the features in a feature structure can be coded by their position, and thus feature-names are omitted from the linear representation. They can be recovered from the type and the positions of the features in the argument list of the term.

We will later on use two feature structures, A and B, to exemplify the machine instructions and its operation. These structures described below, in Figures 2 and 3, represented as an AVM, as a linear term and as a graph (where a grey node denotes a root).

AVM description:

$$\begin{bmatrix} \mathbf{a} \\ f1: \quad [3] \begin{bmatrix} \mathbf{d1} \end{bmatrix} \\ f3: \quad [3] \end{bmatrix}$$

Linear representation: $a([3]d1,[3])$

Graph representation:

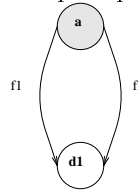

Figure 2: An example feature structure A

The basic operation performed on feature structures is **unification**. There are various definitions for feature structure unification, and we base our unification algorithm (Definition A.17) on Carpenter's definition ([9]). The exact details are given in Appendix A.4. An example of the unification operation is given in figure 4 below, where the arguments are the example feature structures A and B.

AVM representation:                                Graph representation:

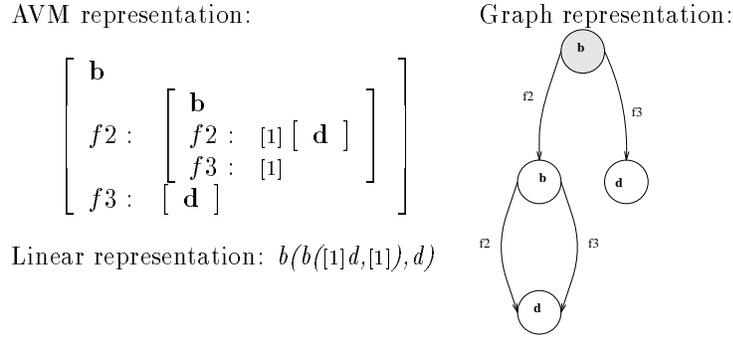

Linear representation: $b(b([1]d,[1]),d)$

Figure 3: An example feature structure B

AVM representation:                                Graph representation:

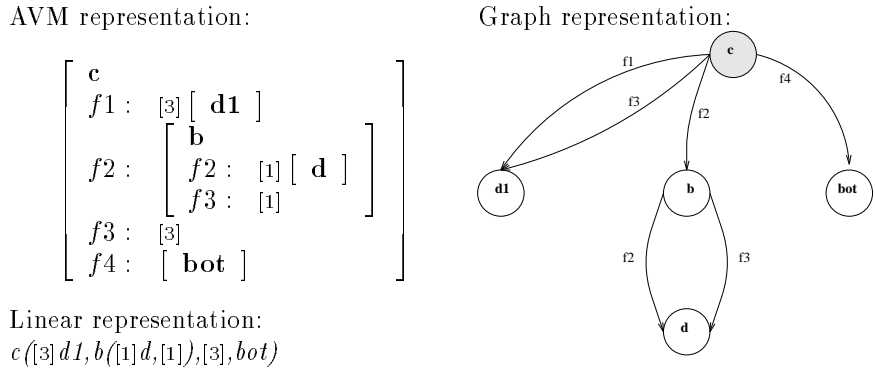

Linear representation:
$c([3]d1,b([1]d,[1]),[3],bot)$

Figure 4: A ⊔ B

## 2.4 Representation of Rules

A **multi-rooted structure** (MRS) is a connected, directed, labeled, finite graph with an ordered non-empty set of distinguished nodes, **roots**. We use MRSs to represent rules, where the graph that is reachable from the *last* root is the **head**[2] of the rule, and those that are reachable from the rest of the roots form the **body** of the rule[3]. A MRS is linearly represented as a sequence of terms, separated by commas, where two occurrences of the same tag, even within two different terms, denotes reentrancy (that is, the scope of the tags is the entire sequence of terms). The head is preceded by '⇒' rather than by a comma. See Appendix A.6 for the exact details.

Application of a rule amounts to unifying its body with a multi-rooted structure resident in memory and producing its head as a result. Since the head and the body of the rule might share values, unifying the body with the existing MRS might influence the head as well. Thus, the head that is produced as a result of the rule application depends also on the resident MRS. The formal definition of rule application is given in Appendix A.6.

An example of a rule, $\rho$, is graphically depicted in figure 5. In this example the rule consists of a MRS of length three, the roots of which are the grey nodes displayed as the uppermost nodes. The head of the rule is the rightmost feature structure, while the body consists of the two leftmost ones. Note that the feature structures in the rule share some nodes; for instance, the node whose type is *d1* is common to all three feature structures. A possible linear representation for $\rho$ is:

$$\rho : b(b([2]d,[2]),[4]d1),\ a([4],[4]) \Rightarrow b(b([2],[4]),d2)$$

---

[2]This meaning of *head* must not be confused with the linguistic definition, referring to the core features of a phrase.
[3]Notice that the intuitive direction is reversed: this will ease later processing of the rules.

In figure 6 a MRS $\sigma$, consisting of two feature structures, is described. We might represent $\sigma$ linearly as:

$$\sigma : a([3]d1,[3]),\ b(b([1]d,[1]),[3])$$

When rule $\rho$ is applied to $\sigma$, $\rho$'s body is unified with $\sigma$. In figure 7 we list the MRS consisting of the new, unified body and the modified head.

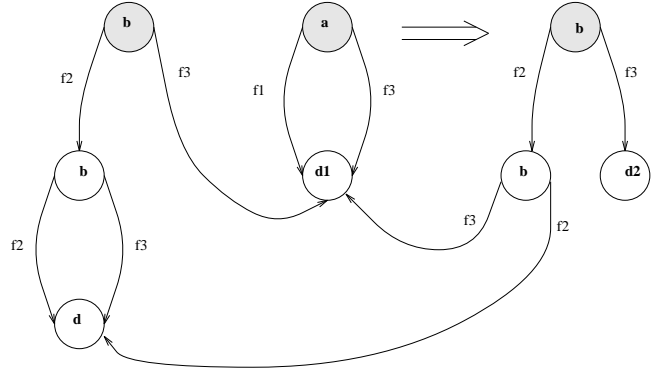

Figure 5: Graphical representation of the rule $\rho$

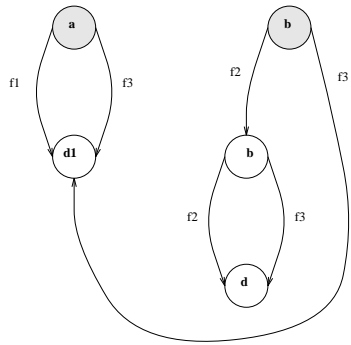

Figure 6: A multi-rooted structure $\sigma$

## 3  The Basic Machine

The heart of the machine design is concentrated on unifying two feature structures. Following the WAM we call one of them *program* and the other – *query*. Both the program and the query are compiled. The program is compiled once, just like an ordinary program, to produce machine instructions. One program is usually designed to be executed against many different queries. Each query is compiled before its execution; the resulting code is executed prior to the execution of the compiled program.

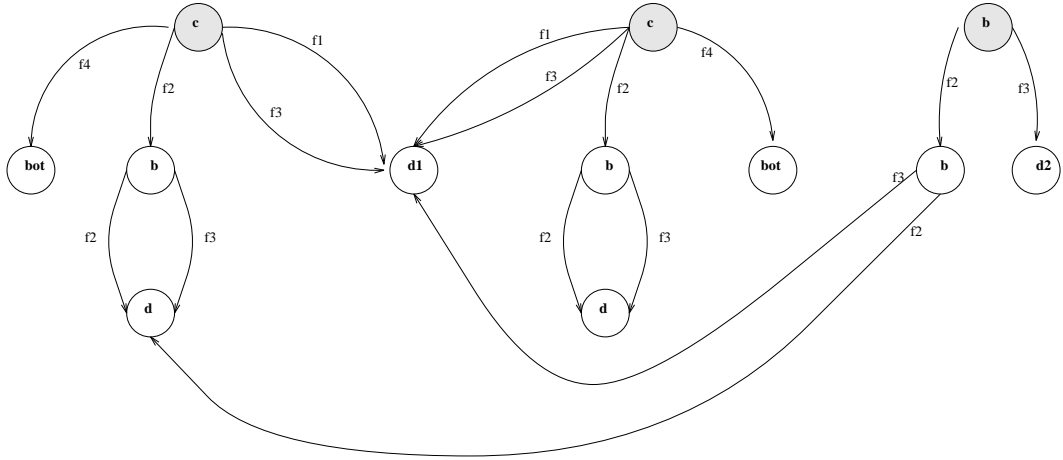

Figure 7: The modified MRS

Processing of a 'query' is aimed towards building a graph representation of the query in the machine's memory. The processing of a 'program' must produce code that, during run-time, will unify the program with a query already resident in memory. The result of the unification will be a new feature structure, represented as a graph in the machine's memory.

## 3.1 First-Order Terms and Feature Structures

While TFSs resemble FOTs in many aspects, it is important to note the differences between them. First, TFSs are typed, as opposed to (ordinary) FOTs. Types can be captured as labeling the nodes of a feature structure. In addition, TFSs label the arcs by feature names, whereas FOTs use a positional encoding for argument structure. A more important difference is the ability to share values within TFSs: while FOTs are essentially trees, with possibly shared leaves, TFSs are directed graphs, i.e., variables can occur anywhere within a feature structure. Moreover, our system doesn't rule out cyclic structures, so that infinite terms can be represented, too.

FOTs are said to be consistent only if they have the same functor and the same arity. TFSs, on the contrary, can be unified even if their types differ (as long as they have a non-degenerate least upper bound). Moreover, their arity can differ, and the arity of the unification result can be greater than the arity of any of the unificands.

These differences are the reasons for many diversions from the original WAM that were necessary in our design. In the following sections we try to emphasize the points where such diversions were made.

## 3.2 Representation of Feature Structures in Memory

Following the WAM, we use a global, one-dimensional array called `HEAP` of data cells, where a cell's address is its index in the array. A global register `H` points to the (current) top element of `HEAP`. Data cells are tagged: STR cells correspond to nodes, while REF cells correspond to arcs. Hence, an STR cell contains the type associated with the node it stands for, and a REF cell contains the address of the node that is the target of the arc it stands for. As we use a total typing system, the arity of a type is the number of arcs leaving a
node of that type. This number is constant for all feature structures of this type; hence, we
can keep the WAM's convention of storing all the outgoing arcs from a node consecutively
following the representation of the node. Given a type and a feature name, we can statically
determine the position of the arc corresponding to this feature is a specific feature structure
of the given type; the subgraph that is the value of this feature can be accessed in one step.
This is a major difference between our method and the approach presented in [3]; we believe
that it leads to a more efficient system without harming the elegance of the machine design.

It is important to note that STR cells differ from their WAM analogs in that they can
be dereferenced when a type is becoming more specific. In such cases, a chain of REF cells
leads to the dereferenced STR cell. Thus, if a feature structure is modified, only its STR
cell has to be changed in order for all pointers to it to 'feel' the modification automatically.

We keep the WAM's convention of representing an uninstantiated variable as a self-
referential REF cell. Such a variable stands for a feature whose value is temporarily un-
known. This is different from the Prolog definition of uninstantiated variables, as in our
system there is always at least partial information as to the type of a structure.

Each node is represented using one cell, and each arc consumes one cell as well. So
for representing a graph of $n$ nodes and $m$ arcs, $n + m$ cells are needed. Of course, during
unification nodes can become more specific and a chain of REF cells is added to the account,
but the length of such a chain is bounded by the depth of the type hierarchy and dereferencing
cuts it occasionally.

As an example, Figure 8 depicts a possible heap representation of the feature structure
*b(b([1]d,[1]),d)*, starting from address 1.

| 1 | STR | b |
|---|-----|---|
| 2 | REF | 4 |
| 3 | REF | 8 |
| 4 | STR | b |
| 5 | REF | 7 |
| 6 | REF | 7 |
| 7 | STR | d |
| 8 | STR | d |

Figure 8: Heap representation of the feature structure *b(b([1]d,[1]),d)*

### 3.3 Flattening Feature Structures

In order to represent a graph on the heap, its linear representation (as a normal term, see
section A.2) is transformed to a set of "equations", each having a flat format, i.e., no nesting
is allowed. To facilitate this we use a set of *registers* $\{X_i\}$ that store *addresses* of feature
structures in memory. We associate a register $i$ with each tag [j] of a normal term. Let
Reg[$j$] be the register associated with the tag [j]. The algorithm for flattening a linear
representation is given in figure 9. In figure 10 there are examples of the equations sets
corresponding to each of the example feature structures.

### 3.4 Processing of a Query

When processing an equation of the form $X_{i_0} = t(X_{i_1}, X_{i_2}, \ldots)$, representing part of a
query, two different instructions can be generated. The first instruction is put_node t/n,
$X_{i_0}$, where $n$ is the arity of $t$. Then, for every argument $X_{i_j}$, an instruction of the form

```
flatten (fs):
    i ← 1; flatten1 (fs);

flatten1 (fs):
    if fs consists of a tag [j] only, return Reg[j]
    else let fs be [j]t(f_1, f_2, ..., f_n)
        if Reg[j] is not defined
            Reg[j] ← i;
            i ← i + 1;
        for k ← 1 to n do
            j_k ← flatten1(f_k);
        print X_{Reg[j]} = t(X_{Reg[j_1]}, ..., X_{Reg[j_n]});
        return Reg[j];
```

Figure 9: The algorithm for flattening terms

| Linear representation: | Set of equations |
| --- | --- |
| $a($[3]$d1,$[3]$)$ | $X1 = a(X2, X2)$ |
| | $X2 = d1$ |
| $b(b($[1]$d,$[1]$),d)$ | $X1 = b(X2, X3)$ |
| | $X2 = b(X4, X4)$ |
| | $X4 = d$ |
| | $X3 = d$ |

Figure 10: Feature structures as sets of equations

put_arc $X_{i_0}$, $j$, $X_{i_j}$ is generated. put_node creates a representation of a node of type $t$ on top of the heap and stores its address in $X_{i_0}$; it also increments H to leave space for the arcs. put_arc fills this space with REF cells.

In order for put_arc to operate correctly, the registers it uses must be initialized. Since only put_node sets the registers, all put_node instructions must be executed before any put_arc instruction is. Hence, we maintain two separate streams of instructions, one for put_node and one for put_arc, and execute the first completely before moving to the other. This compilation scheme is called for by the cyclic character of feature structures: as explained in [3], the original single-streamed WAM scheme would fail on cyclic terms.

The implementation of the two instructions is given in figure 11. Figure 12 lists the machine code that results from compiling the term $b(b($[1]$d,$[1]$),d)$. When this code is executed (first the put_node instructions, then the put_arc ones), the resulting representation of the feature structure in memory is the one shown above in figure 8.

## 3.5 Processing of a Program

Unlike the WAM, in our framework registers that are set by the execution of a query cannot be helpful when processing a program. The reason is that there is no one-to-one correspondence between the sub-terms of the query and the program, as the arity of the feature structures can be different even when the structures are unifiable. We still use the $X_i$ registers, but (with the exception of $X_1$) their old values are not used during execution of the program.

```
put_node t/n, X_i ≡
    HEAP[H] ← <STR,t>;
    X_i ← H;
    H ← H + n + 1;

put_arc X_i, offset, X_j ≡
    HEAP[X_i+offset] ← <REF, X_j>;
```

Figure 11: The implementation of the put instructions

```
put_node b/2,X1        % X1 = b(
    put_arc X1,1,X2    %        X2,
    put_arc X1,2,X3    %           X3)
put_node b/2,X2        % X2 = b(
    put_arc X2,1,X4    %        X4,
    put_arc X2,2,X4    %           X4)
put_node d/0,X4        % X4 = d
put_node d/0,X3        % X3 = d
```

Figure 12: Compiled code for the query $b(b([1]\ d,[1]),d)$

Three kinds of machine instructions are generated when processing an equation of the form $X_{i_0} = t(X_{i_1},\ldots,X_{i_n})$ that is part of a program term. The first instruction is get_structure t/n, $X_{i_0}$, where $n$ is the arity of $t$. For each argument $X_{i_j}$ of $t$ an instruction of the form unify_variable $X_{i_j}$ is generated if $X_{i_j}$ is first seen; if it was already seen in the current term, unify_value $X_{i_j}$ is generated.

For example, the machine code that results from processing the program $a([3]d1,[3])$ is depicted in figure 13. The implementation of these three instructions is given in figure 14.[4]

```
get_structure a/2, X1    % X1 = a(
unify_variable X2        %        X2,
unify_value X2           %           X2)
get_structure d1/0,X2    % X2 = d1
```

Figure 13: Compiled code for the program $a([3]d1,[3])$.

The heart of the implementation lies in the functions unify and unify_type. These functions perform the actual unification of the two feature structures. The get_structure instruction is generated for a feature structure $fs_p$ of a type t which is associated with a register $X_i$. It matches $fs_p$ against a feature structure $fs_q$ that resides in memory using $X_i$ as a pointer to the address of $fs_q$ on the heap. Since $fs_q$ might have undergone some type inference or previous binding (for example, due to previous unifications caused by other instructions), the value of $X_i$ must first be dereferenced. This is done by the function deref (figure 15) which follows a chain of REF cells until it gets to one that either points to an STR cell or is self-referential. This is the value it returns.

---

[4] We use the operator '*' to refer to the contents of an address or a register. We also use '++' as an 'increment' operator, and '--' as 'decrement'.

```
get_structure t/n,X_i ≡
    addr ← deref(X_i); X_i ← addr;
    case HEAP[addr] of
        <REF,addr>:                              % uninstantiated cell
            HEAP[H] ← <STR,t>;
            bind(addr,H);                        % HEAP[addr] ← <REF,H>
            H ← H+1;
            for j ← 1 to n do
                HEAP[H] ← <REF,H>
                enqueue(Q,copy,H);
                H ← H + 1;
        <STR,t'>:                                % a node
            if (unify_type[t,t'](addr) = fail) then fail;

unify_variable X_i ≡
    <action,addr> ← dequeue(Q);
    X_i ← addr;

unify_value X_i ≡
    <action,addr> ← dequeue(Q);
    case action of
        copy:   HEAP[addr] ← *(X_i);
        unify:  if (unify(addr,X_i) = fail) then fail;
```

Figure 14: Implementation of the get/unify instructions

```
function deref(a:address) :  address;
 begin
    <tag,value> ← HEAP[a];
    while (tag = REF and value ≠ a)
        a ← value;
        <tag,value> ← HEAP[a];
    return (a);
end;
```

Figure 15: Implementation of the deref function

The dereferenced value of $X_i$, addr, can either be a self-referential REF cell, i.e., an unbound variable, or an STR cell. In the first case, the feature structure has to be built by the program. A new feature structure is being built on top of the heap (using code similar to that of put_structure) with addr being set to point to it. The second case, in which $X_i$ points to an existing feature structure of type t', is the more interesting one. Here we have to unify an existing feature structure with a new one whose type is t. This is the place to call the compiled code of unify_type with t and t'.

The operation of unify_type will be explained in section 3.6. However, it is important to understand that as a result of this operation a new global queue, Q, is being added to.[5] Every element of Q is a pair <action,addr> where action is either 'copy' or 'unify'. For each feature of the program term, Q determines whether this feature should be simply copied to the unified feature structure, or whether it must be first unified with the appropriate feature in the other structure. The contents of Q will be used by the two unify instructions.

The function unify_type, called by get_structure, is generated as a result of the compilation of the type hierarchy. While the details of this process are given in section 3.6, we list below (figure 16) the compiled code for the unification of our two running examples, of types $a$ and $b$. The second of these structures exists on the heap; the first one has yet to be generated, as it is part of the program, and we don't represent the program as feature structures in memory. Thus the parameters to unify_type, in addition to the two types of the arguments, have to include the address of the existing feature structure in memory.

Note that there isn't *one* unify_type function, but rather many instances of it, arranged as a two-dimensional array indexed by the types of the arguments for the unification. The code we list below, therefore, is referenced as unify_type[a,b](b_addr), where a and b are the types of the arguments, and b_addr is the address of the existing feature structure (of type $b$).

```
unify_type[a,b] (b_addr)
    HEAP[H]   ← <STR,c>;                % since a ⊔ b = c
    HEAP[H+1] ← <REF,H+1>;              % the value of f1 is yet unknown
    enqueue(Q,copy,H+1);
    HEAP[H+2] ← <REF,b_addr+1>;         % f2 is taken from b
    HEAP[H+3] ← <REF,b_addr+2>;         % f3 is taken from b
    enqueue(Q,unify,H+3);               % but still has to be unified with a
    HEAP[H+4] ← <REF,H+5>;              % f4 is a new structure: build it.
    HEAP[H+5] ← <STR,bot>;
    bind(b_addr,H);                     % ≡ HEAP[b_addr] ← <REF,H>
    H ← H + 6;
    return true;
```

Figure 16: Code of unify_type[a,b]

The code for the type unification of the types $a$ and $b$ is rather complex. In many cases the code is much simpler: for example, when the type of the feature structure that is resident in memory is subsumed by the type of the program feature structure, nothing has to be done. As another example, if the program's type is subsumed by the query's type, then the program's additional features have to be added to the resident term. But if no such features exist, the only thing that the function must do is change the type of the resident structure. An example of such a case is depicted in figure 17, where the type unification of the types *d1* and *d* is listed. Note that the function returns 'trivial', rather than 'true', to indicate the fact that no new structures were built in memory. This value will be used by the function **unify** below.

---

[5]We assume the normal operations on queues, where 'enqueue' adds an item on one end and 'dequeue' removes an item from the other end.

```
unify_type[d1,d] (d_addr)
    HEAP[d_addr] ← <STR,d1>;           % since d1 ⊔ d = d1
    return (trivial);
```

Figure 17: Code of `unify_type[d1,d]`

Another example of a trivial case of type unification is that in which the two types are not compatible. The instance of **unify_type** returns 'fail' in such cases. This leads to a call to the function **fail**, which aborts the unification. In section 4.2 we modify the definition of failure and allow some sort of recovery.

In the WAM's original **get_structure** there was no need to call anything like **unify_type**. The WAM's equivalent was a simple check to verify that both structures have the same functor and arity. It is due to the nature of a typed system that a simple equality check has to be replaced by a more complex operation. Since type unification adds information by returning the features of the unified type, this operation builds new structures, in our design, that reflect the added knowledge. Moreover, the WAM's special register S is here replaced by a queue. S was used by the WAM to point to the next sub-term to be matched against. In our design, as the arity of the two terms can differ, there might be a need to hold the addresses of more than one such sub-term. This is what Q is used for – it is being loaded by the various **unify_type** operations with the addresses of all those sub-terms of the program term that have yet to be matched against.

Note that the **unify_variable** instruction resembles very much its WAM analog, in the case of *read mode*. There is no equivalent of the WAM's *write mode* as there are no real variables in our system. However, in **unify_value** there is some similarity to the WAM's modes, where the 'copy' action corresponds to write mode and the 'unify' action to read mode. In this latter case we have to call the function **unify**, just like the WAM does.

The **unify** function (figure 18) is very similar to **unify_type**. In fact, it contains the latter and thus uses it as a subroutine. Recall that **unify_type** is used to perform the type unification of two structures, only one of which is represented on the heap. **unify** does the same, with two differences: first, both feature structures are in memory; second, full unification has to be performed. The first difference is the reason for removing an item from the queue Q and using it as a part of the unification process; the second is realized by recursive calls to **unify** for subgraphs of the unified graphs.

The function **unify** is independent of the types of the structures it operates upon, and thus only one copy of it exists. It receives four parameters: two types and the two addresses of the corresponding feature structures on the heap.

## 3.6 Compilation of the Type Hierarchy

The heart of the unification process lies in the function **unify_type**. This function is generated by compiling the type hierarchy specification; in fact, there are many such functions, one for every pair of types. If the two types are not consistent, their corresponding **unify_type** function simply returns **fail**. However, for consistent types the function produces not only the unified type, but also a feature structure skeleton for this type, that lists all the features that are appropriate for the type. For example, if $t_1$ and $t_2$ are such that $t_1 \sqcup t_2 = t$, then

```
function unify(addr1,addr2:address): boolean;
 begin
    addr1 ← deref(addr1); addr2 ← deref(addr2);
    if (addr1 = addr2) then return(true);
    if (HEAP[addr1] = <REF,addr1>) then
        bind(addr1,addr2); return(true);
    if (HEAP[addr2] = <REF,addr2>) then
        bind(addr2,addr1); return(true);
    H_orig ← H;
    t1 ← *(addr1); t2 ← *(addr2);
    case unify_type[t1,t2](addr2) of
        fail:   return (fail);
        trivial: bind(addr1,addr2); return (true);
    for i ← 1 to arity(t1) + 1 do;
        <action,addr> ← dequeue(Q);
        case action of
            copy:  HEAP[addr] ← <REF,addr1+i>;
            unify: if (not (unify (addr,addr1+i))) then return(fail);
    bind(addr1,H_orig);
    return(true);
end;
```

Figure 18: The code of the **unify** function

the code generated for **unify_type** $t_1, t_2$ will create in memory, when executed, a feature structure of type $t$, including the REF cells for all the appropriate features of $t$.

The generated code for unification of two types assumes that a feature structure of the first type, $t_1$, resides on the heap, while a feature structure of the second type, $t_2$, is part of the program and hence isn't realized on the heap. Thus, for every feature of $t$ that is appropriate for $t_1$ only, execution of the code creates a REF cell that points to the corresponding feature in the first structure. For a feature of $t$ that is appropriate for $t_2$ executing the code creates a self-referential REF cell, but also enqueues to Q the pair <copy,addr> where addr is the address of this feature in memory, thus enabling further processing to copy the future feature to addr. If a feature is appropriate for both $t_1$ and $t_2$, the execution creates a REF cell pointing to the feature of the first structure, and also enqueues a pair <unify,addr> to Q. Later processing will unify the values of this feature in the two structures. Finally, for a feature that is introduced by $t$, execution of the code creates a fresh structure on top of the heap.

In order to generate the **unify_type** functions, the type hierarchy specification has to be processed such that the transitive closure of the subsumption relation is computed. Then, a table is generated in which there is, for every two types, an entry that specifies the least upper bound of these types. Moreover, this table lists also the features of the unified type and their 'origin': whether they are appropriate for $t_1$, $t_2$, both or none of them.

## 4 Extensions

The previous section dealt with a very simple abstract machine, capable of unifying two simple TFSs. We now present two extensions of the machine that will allow processing more complex entities. Section 4.1 describes the representation and application of rules. Section 4.2 details the modifications needed in our design to allow for disjunction within feature structures; the two extensions are orthogonal.

## 4.1 Sequences of Feature Structures

A multi-rooted structure can be thought of as an ordered list of (not necessarily disjoint) feature structures. The unification of such lists goes along the lines of single feature structures unification, and thus our machine doesn't have to be radically modified.

### 4.1.1 Representation of a Multi-rooted Structure

A single feature structure is characterized by one root, whereas a MRS has a sequence of roots. Consequently, a new data structure is needed for storing pointers to the roots. We use an array called ROOTS to store the addresses of the roots of a query MRS. The i-th element ROOTS[i] points to the i-th root of the query. A special purpose register CURR_ROOT is used to index the ROOTS array.

### 4.1.2 Processing of a Query

When processing a query, ROOTS[CURR_ROOT] must hold the address of the root of the feature structure that is currently being built. Recall that each feature structure is converted to a set of equations in turn. The registers are allocated consecutively, so that if, after the execution of the k-th feature structure, the last register that was allocated is $X_i$, then the first register to be allocated for the next feature structure is $X_{i+1}$. We insert a new instruction, advance_q $X_i$ (figure 19), after the code of each feature structure of the query, where $X_i$ is the first register that was allocated for the feature structure.

```
advance_q $X_i$ ≡
    ROOTS[CURR_ROOT] ← $X_i$;
    CURR_ROOT ← CURR_ROOT + 1;
```

Figure 19: The advance_q instruction

Assuming that CURR_ROOT is set to 0 by the initialization of the machine, this modification guarantees that the ROOTS array will store pointers to each feature structure in the query after its execution. To summarize, processing of a query of the form $fs_1, fs_2, \ldots, fs_n$ produces the following code:

```
CURR_ROOT ← 0;
⋮       put_node instructions for $fs_1$
advance_q;
⋮       put_node instructions for $fs_2$
    ⋮
advance_q;
⋮       put_node instructions for $fs_n$
advance_q;
⋮       put_arc instructions for $fs_1, \ldots, fs_n$
```

A technicality arises when a MRS contains a term that consists of a tag only. Such a tag must have appeared in the same MRS earlier, and therefore a register was allocated for it. Therefore, no code is generated for the tag, and two consecutive `advance_q` instructions are created.

Consider, for example, the following query: $a([3]d1,[3]), b(b([1]d,[1]),[3])$. The code that is generated for this query is listed in figure 20.

```
    CURR_ROOT <- 0
put_node a/2,X1         % X1 = a(
    put_arc X1,1,X2     %        X2,
    put_arc X1,2,X2     %            X2)
put_node d1/0,X2        % X2 = d1
advance_q X1
put_node b/2,X3         % X3 = b(
    put_arc X3,1,X4     %        X4,
    put_arc X3,2,X2     %            X2)
put_node b/2,X4         % X4 = b(
    put_arc X4,1,X5     %        X5,
    put_arc X4,2,X5     %            X5)
put_node d/0,X5         % X5 = d
advance_q X3
```

Figure 20: Compiled code for the query $a([3]d1,[3])$, $b(b([1]d,[1]),[3])$

### 4.1.3 Processing of a Program

When processing a program, similar modifications must be made; each feature structure in the MRS is converted to a set of equations in turn. Prior to every code section (corresponding to a single feature structure for which $X_i$ is allocated), an additional new instruction is inserted: `advance_p` $X_i$, whose code is given in figure 21.

> `advance_p` $X_i$ ≡
>     $X_i$ ← ROOTS[CURR_ROOT];
>     CURR_ROOT ← CURR_ROOT + 1;

Figure 21: The `advance_p` instruction

Assuming that `CURR_ROOT` is set to 0 by the initialization of the compilation, this modification guarantees that when `get_structure` will use $X_i$ to match the current program feature structure against a feature structure resident in memory, the address of the resident feature structure is taken from `ROOTS[CURR_ROOT]`, i.e., the k-th program feature structure is guaranteed to be matched against the k-th query feature structure. Here again, if a program contains a feature structure that is reduced to a tag only, then just the new statements that separate consecutive feature structures will be generated, but no special code will be generated for the tag itself.

### 4.1.4 Multi-rooted Structures as Rules

Recall that a rule is represented by a MRS, where the rule's head is the feature structure defined by the last root of the MRS, and the body is the rest of the roots. We demonstrate the manipulation of a (program) rule whose body equals in length to the query.

When compiling a rule, its body generates the same code as that of a program MRS, as explained in the previous section. The head of the rule is treated as a query, and hence the code that is produced for the head is just the code that would have been produced for a single feature structure query of the same form.

For example, consider the following rule: $b(b([2]d,[2]),[4]d1), a([4],[4]) \Rightarrow b(b([2],[4]),d2)$. The generated code for this rule is listed in figure 22. Note that the registers we use for the head are the same set of registers that were used for the body, to accommodate reentrancy.

```
                              % initialization
    CURR_ROOT <- 0
advance_p X1                  % body, first feature structure
get_structure b/2, X1         % X1 = b(
unify_var X2                  %        X2,
unify_var X3                  %            X3)
get_structure b/2, X2         % X2 = b(
unify_var X4                  %        X4,
unify_value X4                %            X4)
get_structure d/0, X4         % X4 = d
get_structure d1/0, X3        % X3 = d1
advance_p X5                  % body, second feature structure
get_structure a/2, X5         % X5 = a(
unify_value X3                %        X3,
unify_value X3                %            X3)
                              % head
    CURR_ROOT <- 0
put_node b/2, X6              % X6 = b(
   put_arc X7                 %        X7,
   put_arc X8                 %            X8)
put_node b/2, X7              % X7 = b(
   put_arc X4                 %        X4,
   put_arc X3                 %            X3)
put_node d2/0, X8             % X8 = d2
advance_q X6
```

Figure 22: Compiled code for the program

## 4.2 Disjunction

Disjunctive values denote indeterminateness: they represent the proposition that more than one value is suitable for some feature. Disjunction within feature structures was discussed in [17, 12, 13, 14] and various kinds of disjunctive values were implemented in different systems. Some complexities arise when *dependent disjunction* is employed, i.e., when the disjunct chosen in one choice point has to correspond with the disjunct chosen in another; the interaction of disjunction with reentrancy is also problematic. For more details, see [14].

As our system is motivated by linguistic considerations, we choose not to maintain the most general notion of disjunctive values. We limit our input language so that only independent disjunctions are allowed, and no disjunct can be reentrant. We augment the syntax of linear terms by adding the '|' operator as a separator between disjuncts; for readability,

we enclose each disjunctive value within curly brackets. We eliminate structure sharing in disjuncts by disallowing tags in the terms that represent them.

Now that disjunction is introduced, the notion of *failure* must be changed: failure no longer entails the abortion of the computation, as failure of one disjunct does not overrule the possibility of another disjunct to succeed. A method of retrying other possibilities must be employed; it is described below.

### 4.2.1 Building Disjunctive Feature Structures

While the syntax of our input language was only slightly modified, a larger number of changes must be made in the design to accommodate disjunction. First, we must show how disjunctive feature structures are built on the heap. When flattening a linear term we treat the '|' operator somewhat like a special type, 'OR', the arity of which is the number of disjuncts. Thus, the term *a({b(bot,d) | a(bot,d1)},d1)* is transformed to the sequence of equations presented in figure 23.

$$
\begin{aligned}
X1 &= a(X2, X3) \\
X2 &= OR(X4, X5) \\
X4 &= b(X6, X7) \\
X6 &= bot \\
X7 &= d \\
X5 &= a(X8, X9) \\
X8 &= bot \\
X9 &= d1 \\
X3 &= d1
\end{aligned}
$$

Figure 23: The disjunctive term *a({b(bot,d) | a(bot,d1)},d1)* as a set of equations

A new instruction, `put_disj`, is introduced; its code is given in figure 24. It creates a special type of cell on the heap: OR-cell, containing the number of disjuncts. For each `put_disj` instruction additional `put_arc` instructions are generated that create a REF cell for each disjunct. The `put_disj` instructions are accumulated in the `put_node` instructions stream. The code that is generated for the above term, taken as a query, is depicted in figure 25, along with the heap after execution of this code.

```
put_disj X_i,n ≡
    HEAP[H] ← <OR,n>;
    X_i ← H;
    H ← H + n + 1;
```

Figure 24: The implementation of the `put_disj` instruction

```
put_node a/2,X1       % X1 = a(
   put_arc X1,1,X2    %        X2,
   put_arc X1,2,X3    %              X3)
put_disj X2,2         % X2 = OR(
   put_arc X2,1,X4    %        X4,
   put_arc X2,2,X5    %              X5)
put_node b/2,X4       % X4 = b(
   put_arc X4,1,X6    %        X6,
   put_arc X4,2,X7    %              X7)
put_node bot/0,X6     % X6 = bot
put_node d/0,X7       % X7 = d
put_node a/2,X5       % X5 = a(
   put_arc X5,1,X8    %        X8,
   put_arc X5,2,X9    %              X9)
put_node bot/0,X8     % X8 = bot
put_node d1/0,X9      % X9 = d1
put_node d1/0,X3      % X3 = d1
```

| 1  | STR | a   |
|----|-----|-----|
| 2  | REF | 4   |
| 3  | REF | 17  |
| 4  | OR  | 2   |
| 5  | REF | 7   |
| 6  | REF | 12  |
| 7  | STR | b   |
| 8  | REF | 10  |
| 9  | REF | 11  |
| 10 | STR | bot |
| 11 | STR | d   |
| 12 | STR | a   |
| 13 | REF | 15  |
| 14 | REF | 16  |
| 15 | STR | bot |
| 16 | STR | d1  |
| 17 | STR | d1  |

Figure 25: Compiled code and heap representation for $a(\{b(bot,d) \mid a(bot,d1)\},d1)$

### 4.2.2 Unifying Disjunctive Feature Structures

Informally, unifying two disjunctive values results in a disjunctive value, in which each disjunct is the result of unifying some disjunct of one unificand with some disjunct of the other. If the number of non-failure results is exactly one, the result is a simple, non-disjunctive, feature structure; if it is zero, the entire unification fails. We can always substitute a non disjunctive value for a disjunction of arity one. The order of the disjuncts is irrelevant. See Appendix A.5 for the formal details.

### 4.2.3 Disjunctive Queries

Consider first the case of disjunctive queries, where the program *does not* contain disjunction. Each of the disjuncts has to be unified, in turn, with the program term. This calls for a major modification in the compiled code of programs: this code is now potentially iterative. Therefore, we encapsulate the code that is generated for each subterm with `loop_start` and `loop_end` instructions. We add a label to every `get_structure` instruction and a corresponding label to every `loop_end`. Using these labels, control can pass from the end of the loop to its beginning and failure recovery can take place, as will be explained below. Figure 26 shows the code that is generated for the (non-disjunctive) program term $a(e([1]d2,[1]),d1)$.

To handle the iteration process we introduce an additional data structure: the *disjunction stack*, DS, stores a *disjunction record* for every disjunction encountered during the execution. Each disjunction record has the following fields:

**start_label** the label corresponding to the `get_structure` instruction against which the disjunction is matched;

**end_label** the label corresponding to the `loop_end` instruction (the instruction to jump to upon failure);

**register** the index of the register that is allocated for the `get_structure` instruction;

**or_addr** the heap address of the OR cell;

**curr_disj** the serial number of the current disjunct;

```
            loop_start X1,l1, l1'
l1:         get_structure a/2, X1    % X1 = a(
            unify_variable X2        %        X2,
            unify_variable X3        %              X3)
            loop_start X2,l2,l2'
l2:         get_structure e/2, X2    % X2 = e(
            unify_variable X4        %        X4,
            unify_value X4           %              X4)
            loop_start X4,l3,l3'
l3:         get_structure d2/0,X4    % X4 = d2
l3':        loop_end l3
l2':        loop_end l2
            loop_start X3,l4,l4'
l4:         get_structure d1/0,X3    % X3 = d1
l4':        loop_end l4
l1':        loop_end l1
```

Figure 26: New code for the program $a(e([1]d2,[1]),d1)$.

**non_fail** the number of disjuncts that successfully passed unification.

A special purpose register D is used to point to the current record of DS.

**loop_start** (figure 27) checks whether the node with which the program is unified is an OR-cell. If it is not, **loop_start** does nothing, so the overhead is minimal. If it is an OR-cell, a new record is added to the disjunction stack. Among the values stored in a disjunction record, the **start_label**, l, should be noted: it uniquely identifies the **get_structure** instruction immediately following the **loop_start** that adds the record; it serves as a way to determine whether or not this **get_structure** instruction is matched against a disjunctive value: if it is, the current disjunction record has l as the value of **start_label**.

**loop_end** receives as a parameter the label l of the corresponding **get_structure** instruction. Therefore, all that **loop_end** has to do in order to know whether or not an OR-cell actually existed is to compare l with the **start_label** of the current disjunction record.

**loop_end** then checks if more disjuncts exist. If so, the register $X_i$, where $i$ is the value of the field **register** of the current record, is set to point to the current disjunct and execution returns to the beginning of the loop. Otherwise, **rearrange_disj** is called: the results of the disjunctive unification are inspected by considering the value of the **non_fail** field in the current disjunction record. If it is zero, the unification fails. If it is one, the OR-cell is eliminated and is replaced by a REF-cell, as the result is non-disjunctive. The REF cells that pointed to the disjuncts are then scanned; those that are self-referential are eliminated. The implementation of the disjunction maintenance auxiliary functions is depicted in figure 28.

How can such a REF cell become self-referential? To understand that, note that the notion of *failure* must be changed. Before disjunction was introduced, **fail** implied the immediate termination of processing. Now, however, failure of one disjunct does not overrule the possibility of successful unification of another. If failure occurs within a disjunction, the next disjunct must be tried. All that has to be done is mark the current disjunct as invalid, by transforming the pointer to it to a self-referential cell. When **loop_end** rearranges OR-cells, it eliminates self-referential cells. The implementation of **fail** is depicted in figure 29.

```
loop_start X_i,l,l' ≡
    addr ← deref(X_i); X_i ← addr;
    if (HEAP[addr] = <OR,k>) then
        add_disj_record (l,l',addr,k,i);
        X_i ← addr + 1;                          % first disjunct

loop_end l ≡
    if (DS[D].start_label = l) then
        DS[D].curr_disj ++;
        if (DS[D].curr_disj < *(DS[D].or_addr)) then
            i ← DS[D].register;                  % more disjuncts left
            X_i ← *(DS[D].or_addr + DS[D].curr_disj);
            jump l;
        else
            rearrange_disj();
```

Figure 27: Implementation of the `loop` instructions

### 4.2.4 Disjunctive Programs

When the program itself is disjunctive one cannot avoid copying the query with which it has to be unified. To understand this recall that, unlike Prolog, our system might return a disjunctive value in cases where one of the unification arguments is disjunctive. Hence, unification has to *collect* possible results, rather than pick a possible value and stick to it until it either fails, in which case another value is chosen, or successfully undergoes unification.

To accommodate disjunctive programs we introduce three new instructions, namely `begin_disj`, `next_disj` and `end_disj`. These instructions are generated such that the code for a disjunctive program term starts with `begin_disj`; prior to each disjunct, including the first, a `next_disj` instruction is generated; and to conclude a disjunctive term we generate `end_disj`. For example, figure 30 shows the code that is generated for the term $a(\{b(bot,d) \mid a(bot,d1)\},d1)$, taken as a program.

To enable copying of the query, we add a field to each disjunction record: `orig_str` stores the address of the original structure that we copy. The implementation of `begin_disj`, given in figure 31, is straightforward: a new disjunction record is added to DS, an OR-cell is built on the heap and the address of the original structure, taken from $X_i$, is recorded. `next_disj` copies the original structure each time it is executed. It also modifies the `end_label` field of the current disjunction record: this field stores the address of the instruction to jump to upon failure. Finally, `end_disj` replaces the original structure with a pointer to the newly-built OR structure and rearranges this OR structure much as `loop_end` does.

When rearranging the OR structure we can check to see if there are nested disjunctions (i.e., if some arc leaving the OR node points to another OR node). We define $\{\{a_1^1|\cdots|a_{n_1}^1\}|\cdots|\{a_1^k|\cdots|a_{n_k}^k\}\}$ to be equivalent to $\{a_1^1|\cdots|a_{n_1}^1|\cdots|a_1^k|\cdots|a_{n_k}^k\}$. Therefore, in such cases the inner disjuncts can be lifted so that all of them reach the same level. This modification is left for an optimization process, not designed yet.

What is left to show is the implementation of the function `copy`. The problem in implementing this function stems from the possibility of cycles in the feature structure to be copied. When scanning it, each node must be labeled, before it is being copied, by the address of its copy. Thus, reentrancy (and directed cycles) can be preserved. There are two alternatives for storing this information: it can either be kept attached to each node, or stored in a temporary table. The first solution implies that each node must now be

```
function add_disj_record (l,l':label, addr:address, n, i:integer):  void
 begin
    D ← D + 1;
    DS[D].start_label ← l;
    DS[D].end_label ← l';
    DS[D].or_addr ← addr;
    DS[D].curr_disj ← 0;
    DS[D].non_fail ← n;
    DS[D].register ← i;
end;

function rearrange_disj ():  void
 begin
    addr ← DS[D].or_addr;
    if (DS[D].non_fail = 0) then
        D ← D - 1;
        fail();
    if (DS[D].non_fail = 1) then           % eliminate OR-cell
        bind (addr,addr+1);                % HEAP[addr] ← <REF,addr+1>
    if (DS[D].non_fail ≠ *(addr)) then     % some disjuncts failed
        i ← 1;
        while (i < *(addr)) do             % for every original disjunct
            t ← addr+i; j ← 1;             % remove self-ref
            while ((HEAP[t] = <REF,t>) and (i < *(addr))) do
                HEAP[t] ← HEAP[t+j];
                j ← j+1; i ← i+1;
            i ← i + 1;
        if (DS[D].non_fail > 1) then
            HEAP[addr] ← <OR,DS[D].non_fail>;    % update OR-cell
    D ← D - 1;
end;
```

Figure 28: Implementation of the disjunction auxiliary functions

```
function fail():  void
 begin
    if (D > 0) then
        addr ← DS[D].or_addr + DS[D].curr_disj;
        DS[D].non_fail --;
        HEAP[addr] ← <REF,addr>;
        jump DS[D].end_label;
    else
        abort;
end;
```

Figure 29: Implementation of the fail function

```
                loop_start X1,l1,l1'
l1:     get_structure a/2, X1          % X1 = a(
        unify_variable X2               %        X2,
        unify_variable X3               %           X3)
        begin_disj X2, 2, ld1           % X2 = {
ld1:    next_disj X4, ld2               %     X4 |
        loop_start X4,l2,l2'
l2:     get_structure b/2, X4           % X4 = b(
        unify_variable X6               %        X6,
        unify_variable X7               %           X7)
        loop_start X6,l3,l3'
l3:     get_structure bot/0, X6         % X6 = bot
l3':    loop_end l3
        loop_start X7,l4,l4'
l4:     get_structure d/0, X7           % X7 = d
l4':    loop_end l4
l2':    loop_end l2
ld2:    next_disj X5, ld3               %         X5
        loop_start X5,l5,l5'
l5:     get_structure a/2, X5           % X5 = a(
        unify_variable X8               %        X8,
        unify_variable X9               %           X9)
        loop_start X8,l6,l6'
l6:     get_structure bot/0, X8         % X8 = bot
l6':    loop_end l6
        loop_start X9,l7,l7'
l7:     get_structure d1/0, X9          % X9 = d1
l7':    loop_end l7
l5':    loop_end l5
ld3:    end_disj                        %              }
        loop_start X3,l8,l8'
l8:     get_structure d1/0, X3          % X3 = d1
l8':    loop_end l8
l1':    loop_end l1
```

Figure 30: Code generated for the program *a({b(bot,d) |a(bot,d1)},d1)*

represented using an extra cell, that can store the address of its copy; the second solution implies the incorporation of a hash table for temporary storing nodes of the feature structure that is currently being copied. Since the copy operation is expected to be performed in other situations (e.g., when manipulating a parser chart), and many graphs are in general expected to be copied, it seems that the better solution is to extend the record representing each node so that a field for an address of its copy is added. The implementation of copy is, thus, straightforward.

## 5 Conclusion

As linguistic formalisms become more rigorous, the necessity of well defined semantics for grammar specifications increases. We presented a first step towards an operational semantics

```
begin_disj X_i,n,l ≡
    add_disj_record (l,1,H,n,i);
    HEAP[H] ← <OR,n>;
    H ← H + n + 1;
    DS[D].orig_str ← deref(X_i);

next_disj X_i,l ≡
    DS[D].curr_disj ++;
    addr ← DS[D].or_addr + DS[D].curr_disj;
    X_i ← addr;
    DS[D].end_label ← l;
    copy (DS[D].orig_str,addr);

end_disj ≡
    bind (DS[D].orig_str,DS[D].or_addr);
    rearrange_disj();
```

Figure 31: Implementation of the disjunction instructions

for such formalisms that are based on typed feature structures. The presentation made use of an abstract machine specifically tailored for this kind of applications. In addition, we described a compiler for a general TFS-based language. The compiled code, in terms of abstract machine instructions, can be interpreted and executed on ordinary hardware. The use of abstract machine techniques is expected to result in highly efficient processing.

This project is still under development. We described here a very simple machine, capable of unifying two feature structures. We then extended the coverage of the machine – and the compiler – by allowing disjunction within feature structures and enabling unification of sequences of feature structures. In other words, our machine is capable of applying a single phrase structure rule. The next step will be the addition of control structures that will enable implementation of a parsing algorithm inherent to the machine. Special constructs will be added to select an appropriate rule out of several possible ones and to maintain temporary results. Future extensions might include negation, list- and set-values and special constructs for generation.

## Acknowledgements

This work is supported by a grant from the Israeli Ministry of Science: "Programming Languages Induced Computational Linguistics". The work of the second author was also partially supported by the Fund for the Promotion of Research in the Technion. We wish to thank Bob Carpenter for his invaluable help during this project.

# A  Theory of Feature Structures

This section gives a brief survey of the theory underlying our design. We follow Carpenter ([7, 9]) in the presentation of the basic building blocks of the TFS theory. The linear representation of terms is based upon Aït-Kaci ([5]). We then give a procedural definition for the unification operation which is parallel to Carpenter's definition. We extend the notion of a feature structure to sequences of feature structures; these sequences will be used for representing phrasal signs and rules.

## A.1  Types and Feature Structures

For the following discussion we fix non-empty, finite, disjoint sets TYPES and FEATS of types and feature names, respectively. We assume that the set FEATS is totally ordered.

A word concerning partial functions is in order here: we use the symbol '↓' (read: 'is defined') to denote that a partial function is defined for some value and the symbol '↑' (read: 'is not defined') to denote the negation of '↓'. Whenever the comparison operator '=' is applied to the result of an application of a partial function, it is meant that the equation holds iff both sides are defined and equal.

**Definition A.1 (Type Hierarchy)** *A partial order relation $\sqsubseteq$ over TYPES × TYPES is an **inheritance hierarchy** if it is bounded complete, i.e., if every up-bounded subset $T$ of TYPES has a (unique) least upper bound, $\sqcup T$, referred to as the **unification** of the types in $T$.*

*If $t_1 \sqsubseteq t_2$ we say that $t_1$ **subsumes**, or is **more general than**, $t_2$; $t_2$ is a **subtype** of $t_1$.*

*Let $\bot$ be the most general type, i.e., $\bot$ is the least upper bound of the empty set of types. Let $\top$ be the most specific type, i.e., $\top = \sqcup \text{TYPES}$. If $\sqcup T = \top$ we say the $T$ is **inconsistent**. Let $\sqcap T$ be the greatest lower bound of the set $T$.*

**Definition A.2 (Feature Structures)** *A **feature structure** $fs$ is a directed, connected, labeled graph consisting of a finite set of nodes $Q$, a root $\bar{q} \in Q$, a partial function $\delta : Q \times \text{FEATS} \to Q$ specifying the arcs and a total node-typing function $\theta : Q \to \text{TYPES}$.*

The nodes of a feature structure are thus labeled by types while the arcs are labeled by feature names. The root $\bar{q}$ is a distinguished node from which all other nodes are reachable. We say that a feature structure is of type $t$ when $\theta(\bar{q}) = t$.

Let FS be the collection of all feature structures over the given FEATS and TYPES.

We use $fs$ (with or without tags, subscripts etc.) to refer to feature structures. We use $Q, \bar{q}, \delta$ and $\theta$ (with the same tags or subscripts) to refer to constituents of feature structures.

Note that all feature structures are, by definition, graphs. Some grammatical formalisms used to have a special kind of feature structures, namely *atoms*; atoms are represented in our framework as nodes with no outgoing edges. For a discussion regarding the implications of such an approach, refer to [9, Chapter 8].

**Definition A.3 (Appropriateness)** *An **appropriateness specification** over the type inheritance hierarchy and the set FEATS is a partial function $Approp : \text{FEATS} \times \text{TYPES} \to \text{TYPES}$, such that:*

- *let $T_f = \{t \in \text{TYPES} \mid Approp(f,t) \downarrow\}$; then for every $f \in \text{FEATS}$, $T_f \neq \phi$ and $\sqcap T_f \in T_f$.*

- *if $Approp(f,t_1) \downarrow$ and $t_1 \sqsubseteq t_2$ then $Approp(f,t_2) \downarrow$ and $Approp(f,t_1) \sqsubseteq Approp(f,t_2)$.*

i.e., every feature is introduced by some most general type, and is appropriate for all its subtypes; and if the appropriate type for a feature in $t_1$ is some type $t$, then the appropriate types of the same feature in $t_2$, which is a subtype of $t_1$, must be at least as specific as $t$.

If $Approp(f,t) \downarrow$ we say that $f$ is appropriate for $t$ and that $Approp(f,t)$ is the appropriate type for the feature f in the type t. We assume that the set of features appropriate for some type is ordered (recall that FEATS is ordered).

**Definition A.4 (Well-typed feature structures)** *A feature structure $(Q, \bar{q}, \delta, \theta)$ is* **well typed** *iff for all $f \in$ FEATS and $q \in Q$, if $\delta(q,f) \downarrow$ then $Approp(f, \theta(q)) \downarrow$ and $Approp(f, \theta(q))$ $\sqsubseteq \theta(\delta(q,f))$.*

i.e., if an arc labeled $f$ connects two nodes, then $f$ is appropriate for the type of the source node; and the appropriate type for $f$ in the type of the source node subsumes the type of target node.

**Definition A.5 (Total well-typedness)** *A feature structure is* **totally well-typed** *iff it is well typed and for all $f \in$ FEATS and $q \in Q$, if $Approp(f, \theta(q)) \downarrow$ then $\delta(q,f) \downarrow$,* i.e., every feature which is appropriate for the type labeling some node must imply the existence of an outgoing arc labeled by this feature.

**Definition A.6 (Appropriateness Loops)** *The appropriateness specification contains a* **loop** *if there exist $t_1, t_2, \ldots, t_n \in$ TYPES such that for every i, $1 \leq i \leq n$, there is a feature $f_i \in$ FEATS such that $Approp(f_i, t_i) = t_{i+1}$, where $t_{n+1} = t_1$.*

**Definition A.7 (Paths)** *A* **path** *is a sequence of feature names, and the set* PATHS $=$ FEATS$^*$ *denotes the collection of paths. The definition of $\delta$ is extended to paths in the natural way:*

$$\delta(q, \epsilon) = q \text{ (where } \epsilon \text{ is the empty path)}$$
$$\delta(q, f\pi) = \delta(\delta(q,f), \pi)$$

**Definition A.8 (Path Values)** *The* **value** *of a path $\pi$ in a feature structure $fs = (Q, \bar{q}, \delta, \theta)$, denoted by $val(fs, \pi)$, is* **non-trivial** *if and only if $\delta(\bar{q}, \pi) \downarrow$, in which case it is a feature structure $fs' = (Q', \bar{q}', \delta', \theta')$, where:*

- $\bar{q}' = \delta(\bar{q}, \pi)$

- $Q' = \{q' \mid \text{there exists a path } \pi' \text{ such that } \delta(\bar{q}', \pi') = q'\}$ *($Q'$ is the set of nodes reachable from $\bar{q}'$)*

- *for every feature $f$ and for every $q' \in Q'$, $\delta'(q', f) = \delta(q', f)$ ($\delta'$ is the restriction of $\delta$ to $Q'$)*

- *for every $q' \in Q'$, $\theta'(q') = \theta(q')$ ($\theta'$ is the restriction of $\theta$ to $Q'$)*

*If $\delta(\bar{q}, \pi) \uparrow$, $val(fs, \pi)$ is defined to be a single node whose type is $\top$.*

**Definition A.9 (Reentrancy)** *A feature structure $fs$ is* **reentrant** *if there exist two non-empty paths $\pi_1, \pi_2$ such that $\delta(\bar{q}, \pi_1) = \delta(\bar{q}, \pi_2)$. In this case the two paths are said to share the same value.*

**Definition A.10 (Subsumption)** $fs_1 = (Q_1, \bar{q}_1, \delta_1, \theta_1)$ **subsumes** $fs_2 = (Q_2, \bar{q}_2, \delta_2, \theta_2)$ *($fs_1 \sqsubseteq fs_2$) iff there exists a total function $h : Q_1 \to Q_2$, called a* **subsumption morphism***, such that*

- $h(\bar{q}_1) = \bar{q}_2$

- *for every $q \in Q_1$, $\theta_1(q) \sqsubseteq \theta_2(h(q))$*

- *for every $q \in Q_1$ and for every $f$ such that $\delta_1(q,f) \downarrow$, $h(\delta_1(q,f)) = \delta_2(h(q), f)$*

i.e., $h$ maps every node in $Q_1$ to a node in $Q_2$ such that the type of the first node subsumes the type of the second, and if an arc labeled $f$ connects $q$ and $q'$ in $Q_1$, then such an arc connects $h(q)$ and $h(q')$ in $Q_2$.

**Definition A.11 (Alphabetic Variants)** *Two feature structures $fs_1$ and $fs_2$ are* **alphabetic variants** *($fs_1 \sim fs_2$) iff $fs_1 \sqsubseteq fs_2$ and $fs_2 \sqsubseteq fs_1$.*

Alphabetic variants have exactly the same structure, and corresponding nodes have the same types. The identities of the nodes are what tell them apart.

## A.2  A Linear Representation of Feature Structures

Representing feature structures as either graphs or attribute-value matrices is cumbersome; we now define a linear representation for feature structures, based upon Aït-Kaci's $\psi$-terms.

**Definition A.12 (Arity)** *The* **arity** *of a type $t$ is the number of features appropriate for it, i.e. $|\{f \mid Approp(f,t) \downarrow\}|$.*

Note that in every totally well-typed feature structure of type $t$ the number of edges leaving the root is exactly the arity of $t$. Consequently, we use the term 'arity' for (totally well-typed) feature structures: the arity of a feature structure of type $t$ is defined to be the arity of $t$.

Let $\{[i] \mid i \text{ is a natural number}\}$ be the set of **tags**.

**Definition A.13 (Terms)** *A* **term** *$\tau$ of type $t$ is an expression of the form $[i]t(\tau_1, \ldots, \tau_n)$ where $[i]$ is a tag, $n \geq 0$ and every $\tau_i$ is a term of some type. If $n = 0$ we sometimes omit the '()'.*

**Definition A.14 (Totally well-typed terms)** *A term $\tau = [i]t(\tau_1, \ldots, \tau_n)$ of type $t$ is* **totally well-typed** *iff:*

- *$t$ is a type of arity $n$;*

- *the appropriate features for the type $t$ are $f_1, \ldots, f_n$, in this order;*

- *for every $i$, $1 \leq i \leq n$, $Approp(f_i, t) \downarrow$;*

- *for every $i$, $1 \leq i \leq n$, if $\tau_i$ is a term of type $t'_i$ and $Approp(f_i, t) = t_i$ then either $t_i \sqsubseteq t'_i$ or $t'_i = \bot$*

We distinguish tags that appear in terms according to the type they are attached to: if a sub-term consists of a tag and the type $\bot$, we say that the tag is **independent**. Otherwise, the tag is **dependent**. We will henceforth consider only terms that are *normal*:

**Definition A.15 (Normal terms)** *A totally well-typed term $\psi = [i]t(\tau_1, \ldots, \tau_n)$ is* **normal** *iff:*

- *$t \neq \top$;*

- *if a tag $[j]$ appears in $\psi$ then its first (leftmost) occurrence might be dependent. If it appears more than once, its other occurrences are independent.*

- *$\tau_1, \ldots, \tau_n$ are normal terms.*

## A.3 Correspondence of Feature Structures and Terms

We are using terms to represent feature structures. We define below an algebra over which terms are to be interpreted. The denotation of a normal term is a totally well-typed feature structure.

**Definition A.16 (Feature structure algebra)** *A* **feature structure algebra** *is a structure* $A = <D_A, \{t_A \mid t \in \text{TYPES}\}, \{f_A \mid f \in \text{FEATS}\}>$, *such that:*

- $D_A$ *is a non-empty set, the* **domain** *of* $A$;
- *for each* $t \in \text{TYPES}$, $t_A \subseteq D_A$ *and, in particular:*
  - $\top_A = \phi$;
  - $\bot_A = D_A$;
  - *if* $t_1 \sqcup t_2 = t$ *then* $t_{1A} \cap t_{2A} = t_A$
- *for each* $f \in \text{FEATS}$, $f_A$ *is a total function* $f_A : D_A \to D_A$

Let $D_G$ be the domain of all typed feature structures over TYPES and FEATS. The interpretation of $t_G$ over this domain is the set of feature structures whose roots have a type $t'$ such that $t \sqsubseteq t'$; the interpretation of $f_G : D_G \to D_G$ is the function that, given a feature structure $fs$, returns $val(fs, f)$.

With each normal term $\psi$ we associate a totally well-typed feature structure $fs$ in the following way:

- if $\psi = \text{[i]}t()$ then $fs = (\{\text{[i]}\}, \text{[i]}, \delta_\uparrow, \theta_t)$ where $\delta_\uparrow$ is undefined for every input and $\theta_t(\text{[i]}) = t$;
- if $\psi = \text{[i]}t(\tau_1, \ldots, \tau_n)$ then $fs = (Q, \text{[i]}, \delta, \theta)$ where $\theta(\text{[i]}) = t$ and for every $j$, if $f_j$ is the $j$-th appropriate feature of the type $t$, then $\delta(\text{[i]}, f_j) = q_j$ and $q_j$ is the root of the feature structure associated with $\tau_j$. $Q$ is $\{\text{[i]}\} \cup \bigcup_j Q_j$ where $Q_j$ is the set of nodes in the feature structure associated with $\tau_j$.

Conversely, with each feature structure $fs = (Q, \bar{q}, \delta, \theta)$ we associate a normal term $\psi = \text{[i]}t(\tau_1, \ldots, \tau_n)$ where:

- $\text{[i]} = \bar{q}$;
- $t = \theta(\bar{q})$;
- $n$ is the number of outgoing edges from $\bar{q}$;
- for every $j$, $1 \leq j \leq n$, $\tau_j$ is the term associated with $\delta(\bar{q}, f_j)$ where $f_j$ is the $j$-th appropriate feature of $t$;
- if the tag [i] occurs elsewhere in $\tau_1, \ldots, \tau_n$, we replace the term that [i] depends on with the term $\bot()$, making this occurrence of [i] independent.

and if a tag [j] occurs more than once in the term thus constructed, we replace all but its first occurrence with $\bot()$.

To summarize, there is a one-to-one correspondence between totally well-typed feature structures and normal terms. In the sequel we use both representations interchangeably.

Note that the tags are only a means of encoding reentrancy in feature structures. Therefore, when displaying a term in which a tag [i] appears just once in a term, we will sometimes omit the tag for the sake of compactness. Then, we sometimes omit the type of independent tags, which are implicitly typed by $\bot$, and display them as tags only.

## A.4 Unification

Let us now abstract away from the identity of specific nodes in feature structures by identifying alphabetic variants. Unification will be defined for representatives of the equivalence classes of all feature structures (with respect to alphabetic variance); its result will again be such an equivalence class representative. We use the term 'unification' to refer to both the operation and its result.

**Definition A.17 (Unification)** *Let $fs_1 = (Q_1, \bar{q}_1, \delta_1, \theta_1)$ and $fs_2 = (Q_2, \bar{q}_2, \delta_2, \theta_2)$ be such that $Q_1 \cap Q_2 = \phi$ (or use alphabetic variants of them for which this condition holds). Let $\approx$ be the least equivalence relation on $Q_1 \cup Q_2$ such that*

- $\bar{q}_1 \approx \bar{q}_2$
- $\delta_1(q_1, f) \approx \delta_2(q_2, f)$ *if both are defined and $q_1 \approx q_2$*

*Let $[q]_\approx$ be the equivalence class of $q$ with respect to $\approx$. The* **unification** *of $fs_1$ and $fs_2$, $fs_1 \sqcup fs_2$, is a new feature structure $fs = (Q, \bar{q}, \delta, \theta)$, defined as follows:*

- *$Q$ is the set of equivalence classes of $Q_1 \cup Q_2$ with respect to $\approx$*
- *$\bar{q} = [\bar{q}_1]_\approx (= [\bar{q}_2]_\approx)$*
- *$\theta(q)$ is the least upper bound of $\{\theta_1(q_1) \mid q_1 \in Q_1 \text{ and } q = [q_1]_\approx\} \cup \{\theta_2(q_2) \mid q_2 \in Q_2 \text{ and } q = [q_2]_\approx\}$*
- *$\delta(q, f) = q'$ if $(\delta_1(q_1, f) = q_1'$ and $q = [q_1]_\approx$ and $q' = [q_1']_\approx)$ or if $(\delta_2(q_2, f) = q_2'$ and $q = [q_2]_\approx$ and $q' = [q_2']_\approx)$*

*We say that the unification* **fails** *if there exists a node $q \in Q$ for which $\theta(q) = \top$.*

**Theorem A.18** *$fs_1 \sqcup fs_2$ is the least upper bound of $fs_1$ and $fs_2$ with respect to subsumption, if an upper bound exists.*
**Proof:**
See [9].

In the following algorithm we assume the existence of an infinite set of nodes from which a unique *new node* can always be drawn.

**Algorithm A.19 (Unification)** *The* **unification** *of $fs_1$ and $fs_2$ is obtained by calling the function 'unify' (Figures 32 and 33) with $\bar{q}_1$ and $\bar{q}_2$ and considering, as the result, the graph whose root was returned by the function and whose nodes are all the nodes reachable from that root.*

The algorithm assumes that both the arguments and the result reside in memory, represented as graphs, so that when given the root, the function can access all other nodes. As a part of its operation the function defines a morphism $h : (Q_1 \cup Q_2) \to Q$ that associates each node of the arguments with a node in the result.

Since $Q, Q_1$ and $Q_2$ are disjoint we use $\delta$ and $\theta$ without subscripts where the appropriate function can be determined by the identity of its arguments.

The algorithm starts by first unifying the roots of the two structures. Unifying two nodes is done by creating a new node, with the unification of the arguments' types as its type, and modifying the function $h$ accordingly (see below). If any of the arguments is not a member of $Q_1 \cup Q_2$, it is replaced by the new node. The outgoing edges of the arguments are then taken care of: those whose labels were unique to one of the arguments only are simply copied to the new node with their values using the function 'copy'. For those whose labels are common to both arguments, the unification is called recursively.

```
unify(q₁,q₂):
    q ← new_node();
    t ← θ(q₁) ⊔ θ(q₂); if t = ⊤ then return fail; else θ(q) ← t;
    if q₁ ∈ (Q₁ ∪ Q₂) then h(q₁)← q; else substitute(q, q₁);
    if q₂ ∈ (Q₁ ∪ Q₂) then h(q₂)← q; else substitute(q, q₂);

    for all outgoing edges e₁ of q₁,
        if e₁ is labeled f₁ and no f₁-labeled edge leaves q₂,
        then create an f₁-labeled outgoing edge in q and set it to point to copy(δ(q₁, f₁));
    for all outgoing edges e₂ of q₂,
        if e₂ is labeled f₂ and no f₂-labeled edge leaves q₁,
        then create an f₂-labeled outgoing edge in q and set it to point to copy(δ(q₂, f₂));

    for all features f such that f-labeled arcs are leaving both q₁ and q₂,
        create an f-labeled edge in q and set it to point to:
            if h(δ(q₁, f)) = q'₁ and h(δ(q₂, f)) = q'₂ then /* h is defined for both! */
                if q'₁ = q'₂ then q'₁ else unify(q'₁, q'₂);
            if h(δ(q₁, f)) = q'₁ and h(δ(q₂, f)) ↑ then
                unify(q'₁, δ(q₂, f));
            if h(δ(q₂, f)) = q'₂ and h(δ(q₁, f)) ↑ then
                unify(q'₂, δ(q₁, f));
            if h(δ(q₁, f)) ↑ and h(δ(q₂, f)) ↑ then
                unify(δ(q₁, f), δ(q₂, f));
    fill(q);
    return q;
```

Figure 32: The unification algorithm

```
substitute(new,old):
    for every q and every f such that δ(q, f) = old, δ(q, f) ← new;
    for every q such that h(q) = old, h(q) ← new;

copy(q):
    if h(q) ↓ then return h(q)
    else let q' be a new node with the type θ(q)
        set h(q) = q';
        for every f-labeled arc that leaves q, create an f-labeled arc in q'
            and set its value to copy(δ(q, f));

fill(q):
    for all features f that are appropriate for θ(q), if δ(q, f) ↑ then
        δ(q, f) ← new_node();
        θ(δ(q, f))← approp(f, θ(q));
        fill(δ(q, f));
```

Figure 33: The unification algorithm – auxiliary functions

The function $h$ associates the arguments' nodes with nodes in the result. It is being redefined during the algorithm whenever a node $q$ that was already mapped to some image in the result is being unified again. In this case, a new node is created and the image of $q$ has to be redefined. The morphism $h$ helps in determining the condition for halting the recursion: if two nodes are being unified and both their images exist and are equal, there is no need in getting on with the recursion.

The calls to 'fill' ensures that if the arguments of the unification are totally well-typed and the appropriate specification contains no loops, the result is also totally well-typed. If these calls are eliminated, the algorithm returns the unification of its arguments as defined above.

The following propositions will help in understanding the algorithm.

**Proposition A.20** *The function h, defined by the algorithm, is total.*
**Proof:**
*The unification algorithm starts with $fs_1$ and $fs_2$ and defines h for their roots. Then the daughters of the roots are scanned, and 'unify' is called recursively with each daughter if h isn't defined for it. Since both $fs_1$ and $fs_2$ are connected, it is guaranteed that h will be defined over the entire domain when the algorithm ends.*

**Proposition A.21** *If $h(q) = q'$ then $\theta(q) \sqsubseteq \theta(q')$.*
**Proof:**
*$h(q)$ is being set by 'unify' when q is unified with some other node to produce $q'$. The type of $q'$ is the least upper bound of the types of q and the other node, hence it is subsumed by the type of q. The same holds when h is being redefined, as the new value of $h(q)$ has a type that is subsumed by the type of the old value. Finally, if h is being set by 'copy', then the type of $h(q)$ equals the type of q.*

**Proposition A.22** *If an f-labeled arc connects nodes u and v in $Q_1$ or in $Q_2$ then such an arc connects nodes $h(u)$ and $h(v)$ in Q.*
**Proof:**
*Immediate from the construction.*

**Theorem A.23** *The result of the unification algorithm is subsumed by both of its arguments.*
**Proof:**
*The morphism h defined by the algorithm was proved by the above propositions to cohere with the subsumption requirements.*

Note that h defines an equivalence relation on $Q_1 \cup Q_2$ which holds for a pair $(q_1, q_2)$ iff $h(q_1) = h(q_2)$. In fact, $h(q)$ is the equivalence class of q with respect to the $\approx$ relation. This relates our algorithm to Definition A.17.

Let us now assume that the algorithm doesn't issue the call to 'fill'. For this modified algorithm, the following propositions hold:

**Proposition A.24** *If $q_1, \ldots, q_n \in Q_1 \cup Q_2$ are such that for every i, $h(q_i) = q$, then $\theta(q) = \sqcup \theta(q_i)$.*
**Proof:**
*Just like the proof of Proposition A.21.*

**Proposition A.25** *For every node $q \in Q$ there exists a node $q' \in Q_1 \cup Q_2$ such that $h(q') = q$.*
**Proof:**
*When a node node q is introduced by 'copy($q'$)', $h(q')$ is set to q. When a new node q is introduced by 'unify', then if either $q_1$ or $q_2$ are members of $Q_1 \cup Q_2$, h is being set for then, and its value is q. Otherwise, h is being re-defined for some nodes and its new value is the new node.*

**Proposition A.26** *If $\delta(q, f) = q'$ for $q, q' \in Q$ then there exist nodes $p, p' \in Q_1 \cup Q_2$ such that $h(p) = q$, $h(p') = q'$ and either $\delta_1(p) = p'$ or $\delta_2(p) = p'$.*
**Proof:**
*It is clear from the construction that arcs are only being added to the result on account of corresponding arcs in one of the unificands.*

**Theorem A.27** *The result of the unification algorithm (without 'fill') is the most general feature structure that is subsumed by the unification arguments.*

**Proof:**
Suppose that $fs$ was returned by $unify(\bar{q}_1, \bar{q}_2)$ and that there exists a feature structure $fs'$ such that $fs_1 \sqsubseteq fs'$ and $fs_2 \sqsubseteq fs'$. Then there exists a subsumption morphism $h': Q_1 \cup Q_2 \to Q'$. Consider the function $h'': Q \to Q'$ defined as follows:

$$h''(q) = h'(p) \text{ where } h(p) = q$$

By the propositions above it is easy to see that $h''$ is a subsumption morphism, i.e., that $fs \sqsubseteq fs'$.

## A.5 Disjunction

We allow feature structures to be **disjunctive**: a disjunctive term is a set of terms, where we usually use ';' to separate elements of the set.

**Definition A.28 (Unification of disjunctive terms)** *Let $\psi_1 = \{\psi_1^1 | \cdots | \psi_1^n\}$ and $\psi_2 = \{\psi_2^1 | \cdots | \psi_2^m\}$ be terms containing disjunction. Then $\psi_1 \sqcup \psi_2 = \{\psi_1^i \sqcup \psi_2^j \mid 1 \leq i \leq n, 1 \leq j \leq m\} \setminus \{\top\}$. If the result is a singleton, we write $\{\psi\}$ as $\psi$; if the result is empty, the unification fails.*

## A.6 Multi-rooted Structures

**Definition A.29 (Multi-rooted Structures)** *A **multi-rooted structure** (MRS) is a pair $<\bar{Q}, G>$ where $G$ is a finite, directed, labeled graph consisting of a set $Q$ of nodes, a partial function $\delta: Q \times \text{FEATS} \to Q$ specifying the arcs and a total function $\theta: Q \to \text{TYPES}$ labeling the nodes, and where $\bar{Q}$ is an ordered, non-empty list of distinguished nodes in $Q$ called **roots**. A certain node $q$ can appear more than once in $\bar{Q}$. $G$ is not necessarily connected, but the union of all the nodes reachable from all the roots in $\bar{Q}$ is required to yield exactly $Q$. The **length** of a MRS is the number of its roots, $|\bar{Q}|$.*

We use $S, R$ (with or without tags, subscripts etc.) to denote MRSs. We use $\theta, \delta, Q$ and $\bar{Q}$ (with the same tags or subscripts) to refer to the constituents of MRSs.

If $<\bar{Q}, G>$ is a MRS and $\bar{q}_i$ is a root in $\bar{Q}$ then $\bar{q}_i$ defines a feature structure in the natural way: this feature structure is $(Q_i, \bar{q}_i, \delta_i, \theta_i)$ where $Q_i$ is the set of nodes reachable from $\bar{q}_i$, $\delta_i$ is the restriction of $\delta$ to $Q_i$ and $\theta_i$ is the restriction of $\theta$ to $Q_i$.

In view of this notion we can refer to a MRS $<\bar{Q}, G>$ as an ordered sequence $<fs_1, fs_2, \ldots, fs_n>$ of (not necessarily disjoint) feature structures, where each root in $\bar{Q}$ is the root of the corresponding feature structure and $<\bar{Q}, G>$ can be determined by $<fs_1, \ldots, fs_n>$. Note that such an ordered list of feature structures is not a sequence in the mathematical sense: removing an element from the list effects the other elements (due to value sharing among elements). Nevertheless, we can think of a MRS as a sequence where a subsequence is obtained by taking a subsequence of the roots and considering only the feature structures they induce. We use the two referencing methods interchangeably in the sequel.

We extend the linear representation of feature structures to MRSs in the natural way, where ',' separates two consecutive structures of the MRS. We also extend the notion of normal terms to MRSs by requiring that only a first occurrence of some tag within the MRS be dependent.

We extend the notion of subsumption to MRSs in the following way:

**Definition A.30 (Subsumption of multi-rooted structures)** *A MRS $<\bar{Q}, G>$ **subsumes** a MRS $<\bar{Q}', G'>$ if $|\bar{Q}| = |\bar{Q}'|$ and there exists a total function $h: Q \to Q'$ such that:*

- *if $q$ is the i-th root in $\bar{Q}$ then $h(q)$ is the i-th root in $\bar{Q}'$;*
- *for every $q \in Q$, $\theta(q) \sqsubseteq \theta'(h(q))$;*
- *for every $q \in Q$ and $f \in \text{FEATS}$, if $\delta(q,f) \downarrow$ then $h(\delta(q,f)) = \delta'(h(q),f)$.*

**Definition A.31 (Unification of multi-rooted structures)** *Let $S_1 =< \bar{Q}_1, G_1 >$ and $S_2 =< \bar{Q}_2, G_2 >$ be MRSs such that $Q_1 \cap Q_2 = \phi$ and $\mid \bar{Q}_1 \mid = \mid \bar{Q}_2 \mid$. Let $\approx$ be the least equivalence relation on $Q_1 \cup Q_2$ such that*

- *for every $i$, $1 \leq i \leq \mid \bar{Q}_1 \mid$, $\bar{q}_{1_i} \approx \bar{q}_{2_i}$*
- *$\delta_1(q_1,f) \approx \delta_2(q_2,f)$ if both are defined and $q_1 \approx q_2$*

*The **unification** of $S_1$ and $S_2$ is a new multi-rooted structure $S =< \bar{Q}, G >$, where $Q \cap (Q_1 \cup Q_2) = \phi$ and $\mid \bar{Q} \mid = \mid \bar{Q}_1 \mid$, defined as follows:*

- *$Q, \delta$ and $\theta$ are defined as in feature structure unification*
- *for every $i$, $1 \leq i \leq \mid \bar{Q} \mid$, $\bar{q}_i$ is the equivalence class of $\bar{q}_{1_i}$ (and of $\bar{q}_{2_i}$)*

An algorithm for the unification of MRSs can be devised on top of the feature structure unification algorithm in the natural way: given two MRSs, unify the feature structures that are defined by the first roots, then by the next roots etc., until all pairs of feature structures were unified. It is easy to see that the order of the feature structure unification is irrelevant.

It is also apparent that the unification of two MRSs is the most general MRS that is more specific than them both, just as is the case with feature structures.

**Definition A.32 (Rules)** *A **rule** is a MRS with a distinguished last element. If $< X_0, \ldots, X_{n-1}, X_n >$ is a MRS then $< X_0, \ldots, X_{n-1} >$ is its **body** and $X_n$ is its **head**. We write such a rule as $< X_0, \ldots, X_{n-1} \Rightarrow X_n >$*

**Definition A.33 (Grammars)** *A **grammar** is a finite set of rules.*

# B  List of Machine Instructions

The following table lists, for quick reference, the machine instructions and functions, accompanied by a reference to the page in the text in which they are described.

| Query processing | | Program processing | |
| --- | --- | --- | --- |
| put_node t/n, $X_i$ | 12 | get_structure t/n, $X_i$ | 13 |
| put_arc $X_i$,offset,$X_j$ | 12 | unify_variable $X_i$ | 13 |
| advance_q $X_i$ | 17 | unify_value $X_i$ | 13 |
| put_disj $X_i$,n | 20 | advance_p $X_i$ | 18 |
| **Disjunction manipulation** | | **Auxiliary functions** | |
| loop_start $X_i$,l,l' | 23 | deref(a):address | 13 |
| loop_end l | 23 | unify(addr1,addr2):boolean | 16 |
| begin_disj $X_i$,n,l | 26 | add_disj_record(l,l',addr,n,i):void | 24 |
| next_disj $X_i$,l | 26 | rearrange_disj():void | 24 |
| end_disj | 26 | fail() | 24 |

# List of Publications - Laboratory for Computational Linguistics

[**LCL 94-1** ] N. Francez: Contrastive Logic (Revised Version of TR # 668).

[**LCL 94-2** ] J. van Eijck and N. Francez: Procedural Dynamic Semantics, Verb-Phrase Ellipsis, and Presupposition.

[**LCL 94-3** ] Y. Winter and N. Francez: Kataphoric Quantification and Generalized Quantifier Absorption.

[**LCL 94-4** ] S. Lappin and N. Francez: E-Type Pronouns, I-Sums, and Donkey Anaphora, March 1994.

[**LCL 94-5** ] U. Ornan: Basic Concepts in "Romanization" of Scripts, March 1994.

[**LCL 94-6** ] The Complexity of Normal Form Rewrite Sequences for Associativity, April 1994.

[**LCL 94-7** ] U. Ornan and M. Katz: "A New Program for Hebrew Index Based on the Phonemic Script", May 1994.

[**LCL 94-8** ] S. Wintner and N. Francez: "Abstract Machine for Typed Feature Structure", July 1994.